%% file: main.tex
\documentclass[a4paper,english]{article}

\usepackage[T1]{fontenc}
\usepackage[latin9]{inputenc}
\usepackage{babel}
\usepackage{array}
\usepackage{float}
\usepackage{multirow}
\usepackage{amsmath}
\usepackage{amssymb}
\usepackage{graphicx}
\usepackage[unicode=true]
 {hyperref}

\makeatletter

\pdfpageheight\paperheight
\pdfpagewidth\paperwidth

\providecommand{\tabularnewline}{\\}
\floatstyle{ruled}
\newfloat{algorithm}{tbp}{loa}
\providecommand{\algorithmname}{Algorithm}
\floatname{algorithm}{\protect\algorithmname}

\newcommand{\lyxaddress}[1]{
	\par {\raggedright #1
	\vspace{1.4em}
	\noindent\par}
}

\@ifundefined{showcaptionsetup}{}{%
 \PassOptionsToPackage{caption=false}{subfig}}
\usepackage{subfig}
\makeatother

\begin{document}
\title{Defense Against Multi-target Trojan Attacks}
\author{Haripriya Harikumar\textsuperscript{1}, Santu Rana\textsuperscript{1},
Kien Do\textsuperscript{1}, Sunil Gupta\textsuperscript{1}, Wei
Zong\textsuperscript{2}, \\
Willy Susilo\textsuperscript{2}, Svetha Venkastesh\textsuperscript{1}}
\maketitle

\lyxaddress{\textsuperscript{1}Applied Artificial Intelligence Institute (A2I2),
Deakin University, Australia,}

\lyxaddress{\emph{\{h.harikumar, santu.rana, k.do, sunil.gupta, svetha.venkatesh\}@deakin.edu.au}}

\lyxaddress{\textsuperscript{2}University of Wollongong, Australia.}

\lyxaddress{\emph{wz630@uowmail.edu.au,wsusilo@uow.edu.au}}
\begin{abstract}
Adversarial attacks on deep learning-based models pose a significant
threat to the current AI infrastructure. Among them, Trojan attacks
are the hardest to defend against. In this paper, we first introduce
a variation of the Badnet kind of attacks that introduces Trojan backdoors
to multiple target classes and allows triggers to be placed anywhere
in the image. The former makes it more potent and the latter makes
it extremely easy to carry out the attack in the physical space. The
state-of-the-art Trojan detection methods fail with this threat model.
To defend against this attack, we first introduce a trigger reverse-engineering
mechanism that uses multiple images to recover a variety of potential
triggers. We then propose a detection mechanism by measuring the transferability
of such recovered triggers. A Trojan trigger will have very high transferability
i.e. they make other images also go to the same class. We study many
practical advantages of our attack method and then demonstrate the
detection performance using a variety of image datasets. The experimental
results show the superior detection performance of our method over
the state-of-the-arts. 
\end{abstract}

\section{Introduction}

\input{intro.tex}

\section{Method}

\input{method.tex}

\section{Experiments}

\input{exp.tex}

\section{Limitations}

\input{limitation.tex}

\section{Conclusion}

\input{conclusion.tex}

{\small{}\bibliographystyle{plain}
\bibliography{ref}
 }{\small\par}
\end{document}

%% file: intro.tex
Deep learning models have been shown to be vulnerable to various kinds
of attacks \cite{goodfellow2014explaining,Ji_etal_17Backdoor,Gu_etal_17Badnets,chen2017targeted,Chan_Ong_19Poison,Moosavi_etal_17Universal,fawzi2018adversarial}.
Among them Trojan attacks \cite{chen2017targeted,Gu_etal_17Badnets,liu2017neural}
are the hardest to defend against as they are very stealthy. It is
carried out by poisoning the model building process. In its simplest
form, training data is poisoned by swelling the target class training
data with data from the other class overlaid with a small trigger
patch. A model trained on such a poisoned dataset behaves expectedly
with pure data but would wrongly predict non-target class as target
class when a test data is poisoned with the same trigger patch. A
small trigger may not cause any issue with other non-Trojan models
or human users, and thus may escape detection until it is able to
cause the intended harm. Due to its ability to hide during standard
model testing, detection of Trojan backdoors require specialised testing.
\begin{figure}
\begin{centering}
\includegraphics[width=0.6\columnwidth]{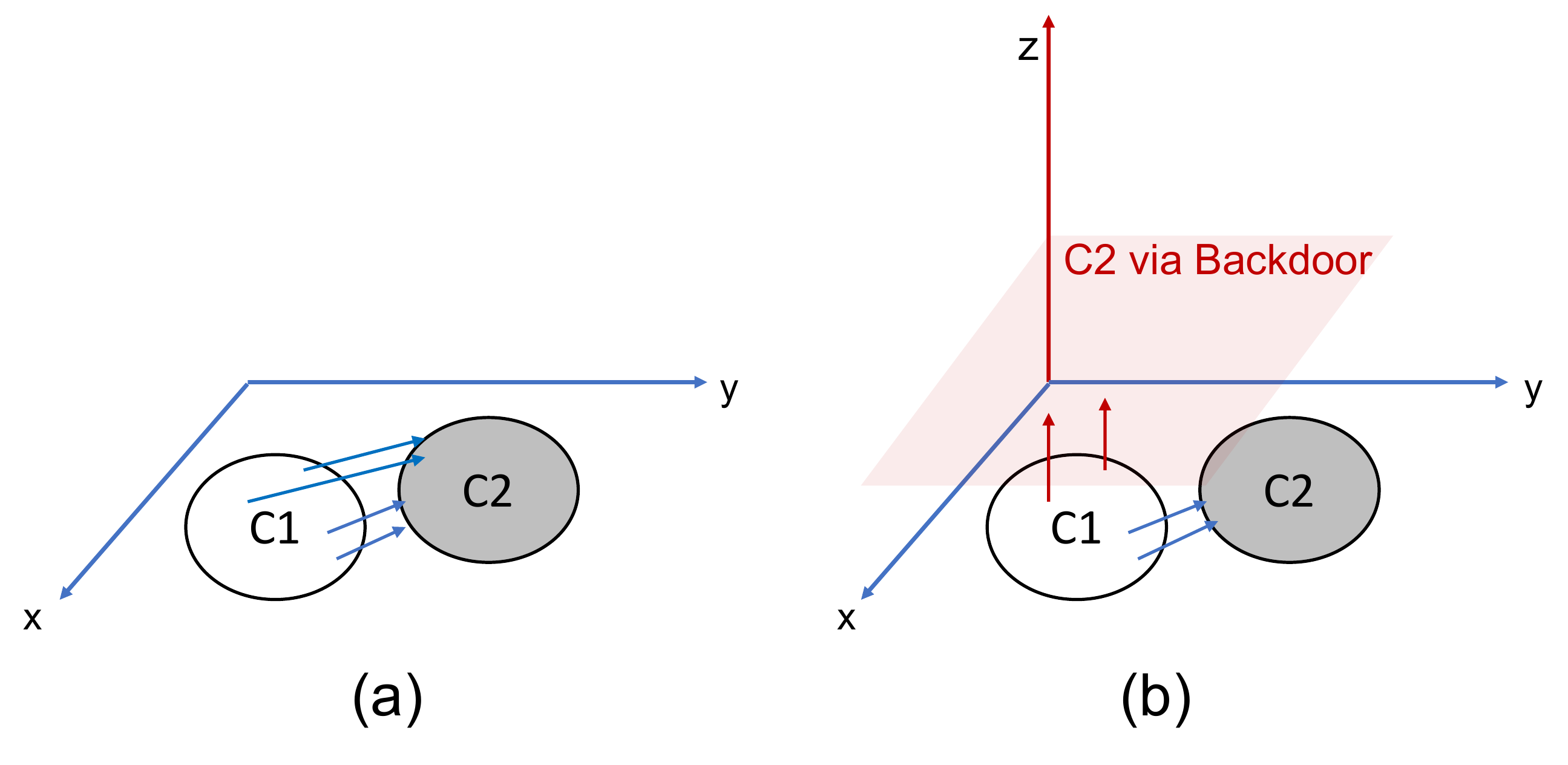}
\par\end{centering}
\caption{(a) Pure model when backdoor is not present. Each image from class
C1 needs different perturbations (blue arrows) to be classified as
class C2. (b) When backdoor is present it creates a shortcut subspace
(red plane). Some of the images from C1 will find perturbations that
are now aligned with this backdoor subspace (red arrows) and thus
are Trojans. All the Trojan triggers are similar and they can take
any image from C1 to the C2 via the backdoor. Some images of C1 will
still find image-specific perturbations (blue arrows) because they
are closer to the original subspace of C2 (gray) than the backdoor
subspace. }
\label{fig:teaser}\vspace{-12bp}
\end{figure}

Testing methods to detect Trojan attacks rely heavily on the assumed
threat model. For Badnet \cite{Gu_etal_17Badnets,gu2019badnets},
where there is only one target class with a trigger that is always
positioned at a fixed location, Neural Cleanse \cite{Wang_etal_19Neural}
and GangSweep \cite{zhu2020gangsweep} provide solid defence mechanisms
through trigger reverse engineering. GangSweep is a slightly more
general version of Neural Cleanse in a sense that it aims to discover
the whole trigger distribution instead of just one trigger. Both these
methods rely on detecting anomalous patterns from the list of reverse
engineered potential triggers. The primary assumption is that the
manually constructed Trojan triggers are very different from others.
Unsurprisingly, they do not work when a large number of classes are
Trojan as anomaly detection would fail. STRIP \cite{Gao_etal_19Strip}
uses a different mechanism to detect the poisoned images and stops
them before they are evaluated by the classifier. It does that by
determining if an incoming image has a signal (trigger) that remains
intact under interpolation in the pure image space (e.g., averaging
the incoming image with known pure images). But it requires the trigger
to be positioned away from the main object in the image such that
the trigger does not get dithered too much during interpolation. It
is thus easy to bypass this defense by carefully positioning the trigger.
\emph{Hence, a proper testing method for multi-target Trojan attack
with no restriction on the positioning of triggers is still an open
problem.} We focus on fixed trigger-based Trojans because it is much
more robust and physically realisable than the more recently input-aware
attacks \cite{nguyen2020input}. Input-aware attacks generate perturbations
for each individual images and thus in applications like autonomous
car where a sensor makes multiple measurements of the same object
at slightly different pose, the input-aware attacks would likely fail
to influence the composite classification process and thus, in our
opinion, they do not pose a significant threat. 

Our objective is to create a defense for multi-target Trojan attacks,
with minimal assumptions about the trigger, e.g., the trigger can
be placed anywhere. Our proposed method is built on trigger reverse
engineering but designed in a way to detect multi-target Trojan attacks.
The intuition of our method is illustrated in Fig \ref{fig:teaser}
through an understanding of the classification surface in pure and
Trojan models. In pure models, each image from class C1 generally
needs different perturbations (blue arrows) to be classified as class
C2 (Fig 1a). But in Trojan models, a backdoor is present in that it
creates a shortcut subspace, shown as red shaded plane (e.g., z =
1) in Fig 1b. Some of the images from C1 will find perturbations that
are now aligned with this backdoor subspace (red arrows) and thus
are Trojan triggers. These Trojan triggers are similar and thus transferable
because they will make any image from C1 to go to C2 through the backdoor
subspace. For some images however, this backdoor perturbations are
larger than directly going from C1 to C2, and thus their perturbation
will remain image-specific (blue arrows in Fig 1b). Our method is
based on finding the transferable perturbations in a given model because
their existence indicates presence of Trojan. We do this in two steps:
perform trigger reverse engineering and then verify their transferability.
To perform trigger reverse engineering, we solve an optimisation problem
to find a small perturbation that takes an image to each of the other
classes. Thus for a 10 class problem, each image will generate 9 triggers.
Once the triggers are identified using a set of pure images (\emph{Data\_Trigger}),
we test them for transferability on another set of images (\emph{Data\_Transfer}).
Each trigger is pasted on the images from \emph{Data\_Transfer}, and
the entropy of the resulting class distribution measured for each
trigger. A Trojan trigger would result in most images going to the
same class, thus resulting in a skewed class distribution and thus
producing small entropy. We provide a way to compute the entropy threshold
below which a perturbation can be termed a Trojan trigger. We call
our proposed attack as Multi-Target Trojan Attack (MTTA) and the associated
defense mechanism as Multi-Target Defense (MTD).

To some extent our trigger reverse engineering process is similar
to GangSweep, but instead of trying to learn a GAN \cite{goodfellow2014generative,goodfellow2020generative}
we use individual triggers straightaway in the detection process.
Because we check for Trojan in each class individually, our method
works even when all the classes are Trojan. We show the efficacy of
our method on four image datasets (MNIST, CIFAR-10, GTSRB, and Youtube
Face). Additionally, we also show that our proposed attack is more
robust than the Badnet and input-aware attacks. Code of the proposed
defense mechanism MTD is provided in the link \href{https://drive.google.com/drive/folders/1lzlFq5fsCtvt6MnLjJ3Ps6SqxGdNYvuT?usp=sharing}{https://bit.ly/3CE1Z3m}.

%% file: method.tex
\begin{figure*}
\centering{}\includegraphics[width=0.9\textwidth]{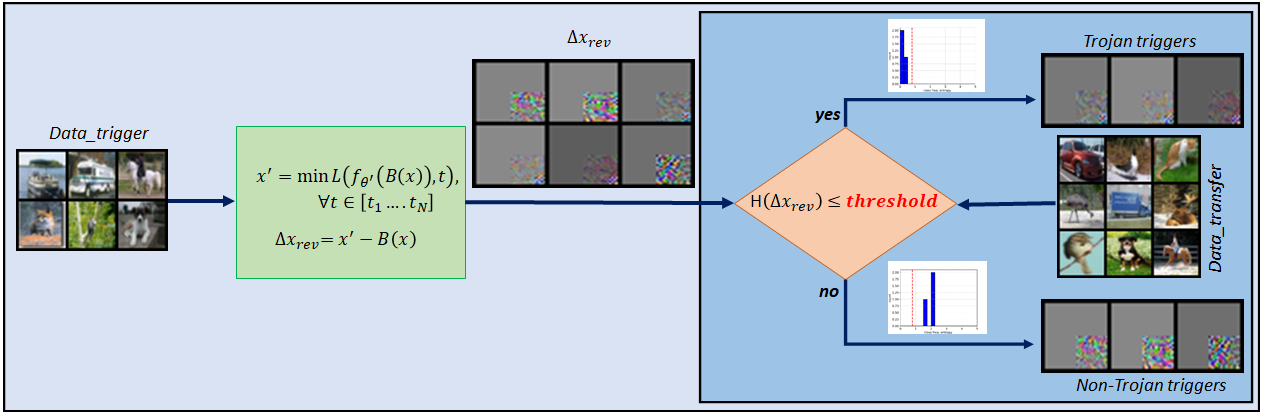}\caption{Schematic diagram of the proposed Multi-target Defense (MTD) method.
Images from the Data\_trigger is used for trigger reverse engineering.
The reverse-engineered triggers are tested on Data\_transfer to check
their transferability. Triggers that produces low entropy for the
class distribution are termed Trojan triggers. The dotted red line
in the entropy plots separate the Trojan and non-Trojan triggers.
The original trigger used is a checkerboard pattern, thus the Trojan
triggers contains a similar pattern (please zoom in to see the pattern).\label{fig:method}}
\end{figure*}
A Deep Neural Network (DNN) can be defined as a parameterized function
$f_{\theta}:\mathcal{X}\rightarrow\mathbb{R}^{C}$ that recognises
class label probability of an image $x\in\mathcal{X}$ where $\mathcal{X}\sim P_{\mathcal{X}}$
and $\theta$ represents the learned function's parameter. The image
$x$ will be predicted to belong to one of the $C$ classes. The output
of the DNN is a probability distribution $p\in\mathbb{R}^{C}$ over
$C$ classes. Let us consider probability vector for the image $x$
by the function $f_{\theta}$ as $\left[p_{1}....p_{C}\right]$, thus
the class corresponding to $x$ will be $\text{argmax}_{i\in[1..C]}p_{i}$.
DNN will learn its function parameters, weights and biases with the
training dataset, $\mathcal{D}_{train}=\left\{ \left(x_{i},y_{i}\right)\right\} _{i=1}^{M}$,
where $M$ is the number of data in the training set and $y_{i}$
is the ground-truth label for the instance $x_{i}$. Next, we discuss
the threat model settings and the defense mechanism.

\subsection{Threat Model}

The attack setting we introduce has three key elements: 1) multiple
triggers, $\left[\Delta x_{1}....\Delta x_{N}\right]$; 2) multiple
target classes, $\left[t_{1}....t_{N}\right]$; and 3) trigger can
be placed anywhere in the image. We use a square patch as trigger
which when put on the image cause misclassification. However, the
attacker can use triggers of any shape as long as it is not covering
a large part of the whole image. We have different triggers associated
with each target class. The target classes are a subset of classes
randomly chosen from the known set of classes of the dataset i.e.
$N<C$. The practicality of this trigger anywhere lies in the fact
that in the real world an attacker can put a sticker on any location
of the image, instead of carefully positioning it like Badnet. This
sticker can be opaque or semi-transparent.

Mathematically, for Trojan model the original DNN model with model
parameters $\theta$ will be replaced by Trojan model parameters $\theta^{'}$
denoted as $f_{\theta'}(.)$. The pure input image is perturbed by
a trigger which is of size comparatively lesser than the original
image size. The following shows the composition method for the trigger
and the images.\newline

\textbf{\emph{Definition 1}} : A trigger is formally defined as a
small square image patch $\Delta x$ of size $s$ that is either physically
or digitally overlaid onto the input image $x$ at a location $(i^{*},j^{*})$
to create a modified image $x'$. Concretely, an image of index $k$
of the dataset $x_{k}$ is altered into an image $x'_{k}$ by,\vspace{-12bp}
\begin{equation}
x'_{k}(i,j)=\begin{cases}
(1-\alpha(i^{'},j^{'}))x_{k}(i,j) & \textrm{if}\:i\in[i^{*},i^{*}+s],\\
\quad+\alpha(i^{'},j^{'})\Delta x(i^{'},j^{'}) & j\in[j^{*},j^{*}+s]\\
\\
x_{k}(i,j) & \textrm{elsewhere}
\end{cases},\label{eq:ImageCorruption}
\end{equation}
\noindent where $(i^{'},j^{'})$ denote the local location on the
patch $(i',j')=(i-i^{*},j-j^{*})$ as defined in \cite{harikumar2021scalable}.
The transparency of the trigger is controlled by the weight, $\alpha$.
This parameter can be considered as a part of the trigger, and we
will be inclusively mentioning it as $\Delta x$. Meanwhile, the rest
of the image is kept the same. In our setting, $(i^{*},j^{*})$ can
be at any place as long as the trigger stays fully inside the image.

\subsection{Trojan Detection}

We use the validation dataset of pure images for trigger reverse engineering
and transferable trigger detection by splitting it into two separate
datasets: a) \emph{Data\_Trigger} - fro trigger reverse engineering,
and b) \emph{Data\_Transfer} - for checking transferability of the
reverse engineered triggers. For each image in the \emph{Data\_Trigger}
we find a set of perturbations by setting each class as a target class.
We restrict the search space of trigger reverse engineering by using
a mask that spans no more than 1/4th the size of the image. Here we
arbitrarily assume that the trigger is not larger than 1/4th of an
image, a sensible assumption if we need to consider the stealth requirement
of a Trojan trigger. Our framework is not constrained by this assumption,
although optimisation efficiency may vary. Each reverse engineered
trigger is then used on the images of \emph{Data\_Transfer} to compute
the class-distribution entropies. If a perturbation is the Trojan
trigger, then it will transfer to all the images and the class distribution
would be peaky at the target class, resulting in a small entropy value.
We provide a way to compute the entropy threshold below which a perturbation
is termed Trojan trigger. Below we provide the details of each steps.
\begin{algorithm}
\textbf{Inputs }: $x$, $C$, $f_{\theta'}(.)$, $x_{test}$, threshold

\textbf{Outputs}: target\_classes,\textbf{ Boolean }trojan\_model 

\textbf{for} each \emph{class} in $C$ \textbf{do}

~~~~~Compute optimised image, $x'$ with \emph{class} using Eq
\ref{eq:2}.

~~~~~Compute reverse engineered trigger, $\Delta x_{rev}$ with
Eq \ref{eq:3}.

~~~~~Compute entropy, $\text{H}(\Delta x_{rev})$ using $x_{test}$
with Eq \ref{eq:5}.

~~~~~\textbf{if} $\left(\text{H}(\Delta x_{rev})\right)\leq\text{threshold}$
\textbf{then}

~~~~~~~~~~~~~target\_classes.append(\emph{class})

~~~~~\textbf{end if} 

\textbf{end for}

\textbf{if} $\text{length}\left(\text{target\_classes}\right)\geq\text{1}$
\textbf{then}

~~~~~ trojan\_model $=$ \textbf{True}

\textbf{else}

~~~~~ trojan\_model $=$ \textbf{False}

\textbf{end if}

\caption{Multi Target Defense (MTD).\label{alg:Multi-Target-Defense} }

\end{algorithm}

\subsubsection{Trigger Reverse Engineering}

Given an image, $x\in\mathbb{R^{\mathrm{\mathit{Ch\times H\times W}}}}$,
where $\mathrm{\mathit{Ch,H,W}}$ are the number of channels, height,
and width and a target label $y$, we define $B(x)$ as the mask that
only keep inside pixels active for the optimisation i.e., \vspace{-5bp}
\begin{equation}
B(x)=x\odot B,
\end{equation}

\vspace{-5bp}
\noindent where $\odot$ is the element-wise product and B is a binary
matrix. B has a value of 1 across a region $H/4\times W/4$ across
all $Ch$ channels, and can be positioned anywhere, as long as it
is fully within the image. We then minimise the cross-entropy loss
between the predicted label for $B(x)$ and the target label $y$:\vspace{-15bp}

\begin{center}
\begin{equation}
x^{'}=\mathcal{L}\left(f_{\theta'}(B(x)),y\right).\label{eq:2}
\end{equation}
\par\end{center}

\vspace{-11bp}

\begin{flushleft}
The reverse engineered trigger which we denote as $\Delta x_{rev}$
is the difference between $x^{'}$ and $x$:
\par\end{flushleft}

\vspace{-20bp}

\begin{center}
\begin{equation}
\Delta x_{rev}=x^{'}-B(x).\label{eq:3}
\end{equation}
\par\end{center}

\vspace{-14bp}

\subsubsection{Transferability Detection\label{subsec:Entropy-score}}

To check for transferability, we compute the entropy \cite{shannon1948mathematical}
of the class distribution for each reverse engineered trigger when
used on all the images of the \emph{Data\_transfer} as follows,\vspace{-12bp}

\begin{equation}
\text{H}(\Delta x_{rev})=-\sum_{i=1}^{C}p_{i}\text{log}_{2}(p_{i}),\label{eq:5}
\end{equation}

\vspace{-8bp}
\noindent where $\{p_{i}\}$ is class probability for the \emph{i}'th
class for using that perturbation. The entropy of a Trojan trigger
will be zero if the Trojan attack success rate is 100\%. However in
real-world situation, we assume a slightly less success rate that
lead to a non-zero entropy value. The following lemma shows how to
compute an upper bound on the value of this score for the Trojan models
in specific settings, which then can be used as a threshold for detecting
Trojans.\newline \newline
\textbf{\emph{Lemma 1}} :\emph{ Let the accuracy of Trojan model on
data with embedded Trojan triggers to be $(1-\delta),$ where $\delta<<1$,
and let there be $C$ different classes. If $\Delta x_{rev}$ is a
Trojan trigger then the entropy computed by Eq \ref{eq:5} will be
bounded by}\vspace{-13bp}

\begin{equation}
\text{H(\ensuremath{\Delta x_{rev})}}\leq-(1-\delta)*\text{log}_{2}(1-\delta)-\delta*\text{log}_{2}(\frac{\delta}{C-1}).\label{eq:threshold}
\end{equation}
\vspace{-11bp}

The above lemma can easily be proved by observing that the highest
entropy of class distribution in this setting happens when $(1-\delta)$
fraction of the images go to the target class $t^{'}$ and the rest
$\delta$ fraction of the images gets equally distributed in the remaining
$(C-1)$ classes. This entropy score is independent of the type and
size of triggers used and is universally applicable. This threshold
computation has been adopted from STS \cite{harikumar2021scalable}.
The overall algorithm is provided in Algorithm \ref{alg:Multi-Target-Defense}
and a visual sketch of the method is presented in Fig \ref{fig:method}.

%% file: exp.tex
We evaluate our proposed defense method on four datasets namely, MNIST,
German Traffic Sign Recognition Benchmark (GTSRB) \cite{stallkamp2011german},
CIFAR-10 \cite{Krizhevsky09learningmultiple}, and YouTube Face Recognition
(YTF) dataset \cite{ferrari2018extended}. We use Pre-activation Resnet-18
\cite{nguyen2020input} as the classifier for CIFAR-10 and GTSRB and
Resnet-18 \cite{he2016deep} for YTF. We use a simple network architecture
\cite{nguyen2020input} for MNIST dataset. The details of the datasets
and the attack settings are shown in Table \ref{tab:Dataset}. 

We train the Pure and Trojan classifiers using SGD \cite{bottou2012stochastic}
with initial learning rate of 0.1 and used learning rate scheduler
after 100, 150, and 200 epochs, weight decay of 5e-4 and the momentum
of 0.9. We use batch size of 128 for CIFAR-10 and MNIST, and 256 for
GTSRB with the number of epochs as 250. For YTF we use the batch size
of 128 and number of epochs as 50. For Trojan models, the target
and non-target class ratio we have used is 70:30 ratio except for
YTF which is 30:70 as it contains lots of classes and we found it
hard to obtain a good pure accuracy with 70:30 poisoning ratio. While
training the Trojan model, the ratio of Trojan data in a batch for
MNIST and CIFAR-10 is set to 10\% of the batch size 2\% for GTSRB
and 0.2\% for YTF. They are chosen to minimise the impact of Trojan
data on pure data accuracy. 

We use square triggers of sizes 4$\times$4, and 8$\times$8. with
trigger transparency of 1.0. We use random pixel values to create
class-specific triggers. The purpose of random colored triggers are
two-fold: a) to show that attack is potent even when triggers are
not optimally distinct, and b) that the defense works without any
structure in the triggers. For trigger reverse engineering we use
Adam optimizer \cite{Kingma_Ba_14Adam} with a constant learning rate
of 0.01. 

We term the accuracy computed on the pure data on the ground-truth
labels as the \emph{pure accuracy} and the accuracy on the Trojan
data corresponding to the intended target classes as the \emph{Trojan
accuracy}. To demonstrate the strengths of MTTA, we also analyse two
additional properties: a) Robustness - how badly the Trojan accuracy
is affected by i) image translation, to mimic the misplacemet of the
object detection bounding box and ii) trigger translations to mimic
the misplacement during physical overlaying of the trigger on the
image. In both the cases a part of the trigger may get lost; and b)
Invisibility - how well Trojan data can hide from the pure classifiers.
If an attack is visible then it would cause unintended side-effect
by attacking pure classifiers too and thus compromise their stealth.
We believe that strong robustness and complete invisibility are the
hallmark of an extremely potent Trojan attack. 

We compare with STRIP \cite{Gao_etal_19Strip}, FinePruning \cite{liu2018fine}
, STS \cite{harikumar2021scalable} and NAD \cite{li_21neural} Neural
Cleanse . However, STRIP is a test time defense and thus do not admit
the same metric as Neural Cleanse and MTD. We do not compare with
DeepInspect \cite{Chen_etal_19Deepinspect} or GangSweep \cite{zhu2020gangsweep}
as we believe that it would be unreasonable to train a GAN (used in
both) with the small number of trigger reverse-engineering that we
will be performing (20 -128) for different datasets. 

\subsection{Effectiveness of MTTA}

The accuracies of the pure and the MTTA Trojan models are reported
in Table \ref{acc}. The performance shows that across various configuration
choices, the proposed attack strategy succeeds in providing a high
Trojan effectiveness ($\sim$100\%) whilst keeping the pure accuracy
close to the pure model accuracy.
\begin{table}
\centering{}\caption{Pure accuracy of Pure models and MTTA Trojan Models as well as the
Trojan accuracy of the MTTA Trojan models.\label{acc}}
\begin{tabular}{|c|c|c|c|c|c|}
\hline 
\multirow{3}{*}{Dataset} & \multicolumn{3}{c|}{Pure accuracy} & \multicolumn{2}{c|}{Trojan accuracy}\tabularnewline
\cline{2-6} \cline{3-6} \cline{4-6} \cline{5-6} \cline{6-6} 
 & Pure & \multicolumn{2}{c|}{Trigger size} & \multicolumn{2}{c|}{Trigger size}\tabularnewline
\cline{3-6} \cline{4-6} \cline{5-6} \cline{6-6} 
 & model & 4$\times$4 & 8$\times$8 & 4$\times$4 & 8$\times$8\tabularnewline
\hline 
MNIST & 99.53 & 98.83 & 99.24 & 99.76 & 99.98\tabularnewline
\hline 
GTSRB & 98.85 & 98.84 & 100.0 & 100.0 & 100.0\tabularnewline
\hline 
CIFAR10 & 94.55 & 93.93 & 94.39 & 100.0 & 100.0\tabularnewline
\hline 
YTF & 99.70 & 99.55 & 99.34 & 96.73 & 99.79\tabularnewline
\hline 
\end{tabular}
\end{table}

\subsection{Robustness of MTTA }

We use the MTTA Trojan model which was trained on CIFAR-10 with 8$\times$8
Trojan triggers to demonstrate the robustness of MTTA under both slight
misplacement of the image window and the trigger placement. To carry
out the test, we translate the image up or down and pad the added
rows with white pixels. For trigger translation we do the same way
only for the trigger before compositing it with the untranslated images.
\begin{figure*}
\subfloat[Image translate up.\label{fig:Translate-up}]{\begin{centering}
\includegraphics[width=0.24\textwidth]{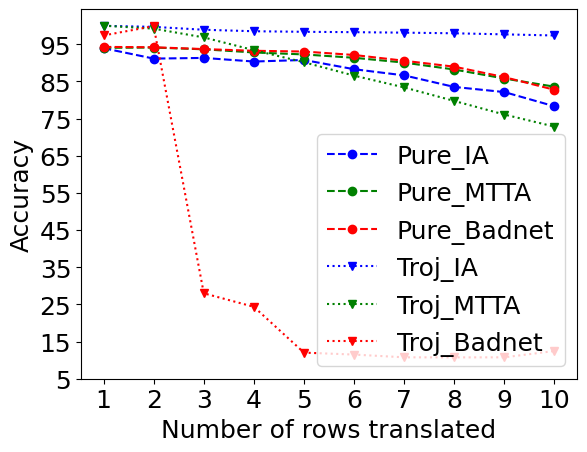}
\par\end{centering}
}\subfloat[Image translate down.\label{fig:Translate-down}]{\begin{centering}
\includegraphics[width=0.24\textwidth]{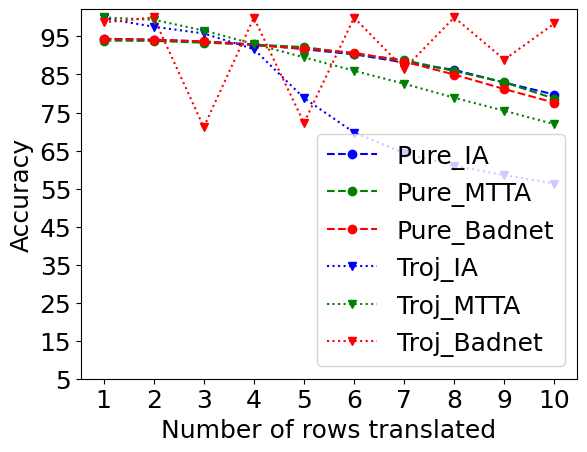}
\par\end{centering}
}\subfloat[Trigger translate up.\label{fig:Translate-right}]{\begin{centering}
\includegraphics[width=0.24\textwidth]{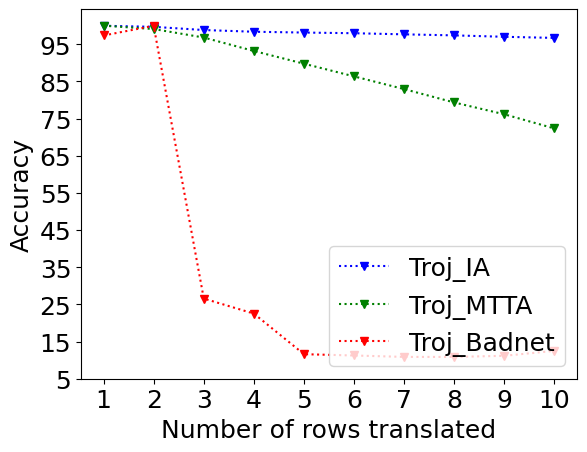}
\par\end{centering}
}\subfloat[Trigger translate down.\label{fig:Translate-left}]{\begin{centering}
\includegraphics[width=0.24\textwidth]{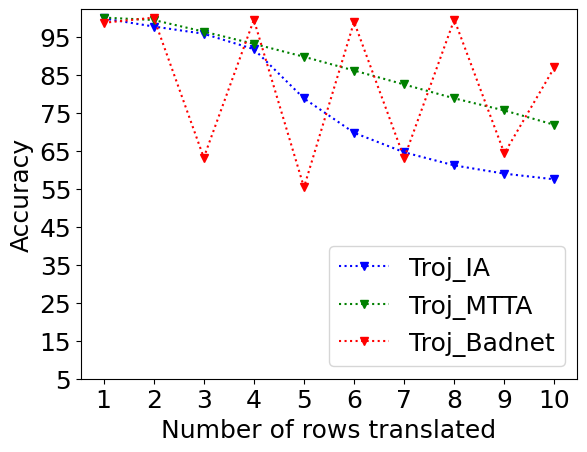}
\par\end{centering}
}\caption{Robustness of MTTA attack against image/trigger translations. Pure
and Trojan (denoted as Troj\_) accuracies vs number of rows translated
for Badnet, Input-aware attack (IA), and MTTA. Figure \ref{fig:Translate-up}
shows the accuracy when we translate the images up, Figure \ref{fig:Translate-down}
when we translate the images down, Figure \ref{fig:Translate-right}
when we translate the triggers up, and Figure \ref{fig:Translate-left}
when we translate the triggers down. Pure accuracies are not afected
by trigger translations, and thus not reported. \label{fig:robuts_mtta}}
\end{figure*}

The plots show the pure and Trojan accuracy when we translate image
up (Figure \ref{fig:Translate-up}), translate image down (Figure
\ref{fig:Translate-down}), translate trigger up (Figure \ref{fig:Translate-right}),
and translate trigger down (Figure \ref{fig:Translate-left}). We
have chosen three attack models, the 8$\times$8 trigger trained CIFAR-10
MTTA model, Input-aware attack (denoted as IA in Figure \ref{fig:robuts_mtta})
and a Badnet trained with a 8$\times$8 checkerboard trigger placed
at the top-right corner. When translating up, we see that Badnet is
disproportionately affected, whilst input-aware attack remained largely
resilient. MTTA also dropped, but only slightly. When translating
down, the Badnet remained resilient as the trigger patch remained
within the translated image, but the Trojan effectiveness (Trojan
accuracies) of input-aware attack dropped way more than MTTA. When
trigger is translated, the image underneath is not affected, and hence
pure accuracies are not affected. But the Trojan effectiveness drops.
However, we see again that whilst MTTA remains more or less resilient,
in one case Badnet dropped catastrophically (Fig \ref{fig:Translate-right})
and in another case input-aware attack dropped way more than MTTA
(Figure \ref{fig:Translate-left}). This shows that MTTA is a robust
attack, especially when carried out in the physical space. Image
and trigger translation based on left and right rows of the images
are reported in supplementary. 

\subsection{Robustness of MTTA against STRIP, FinePruning, STS and NAD}

We have tested CIFAR-10 8$\times$8 trigger trained MTTA Trojan model
against a state-of-the-art test time defense mechanisms such as STRIP
\cite{Gao_etal_19Strip}, FinePruning \cite{liu2018fine} , STS \cite{harikumar2021scalable}
and NAD \cite{li_21neural}. We have 7 target classes (class \emph{6,
9, 0, 2, 4, 3,} and \emph{5}) with 7 different triggers for each target
class.
\begin{figure*}[t]
\centering{}\subfloat[STRIP\label{strip-2}]{\begin{centering}
\includegraphics[width=0.4\textwidth]{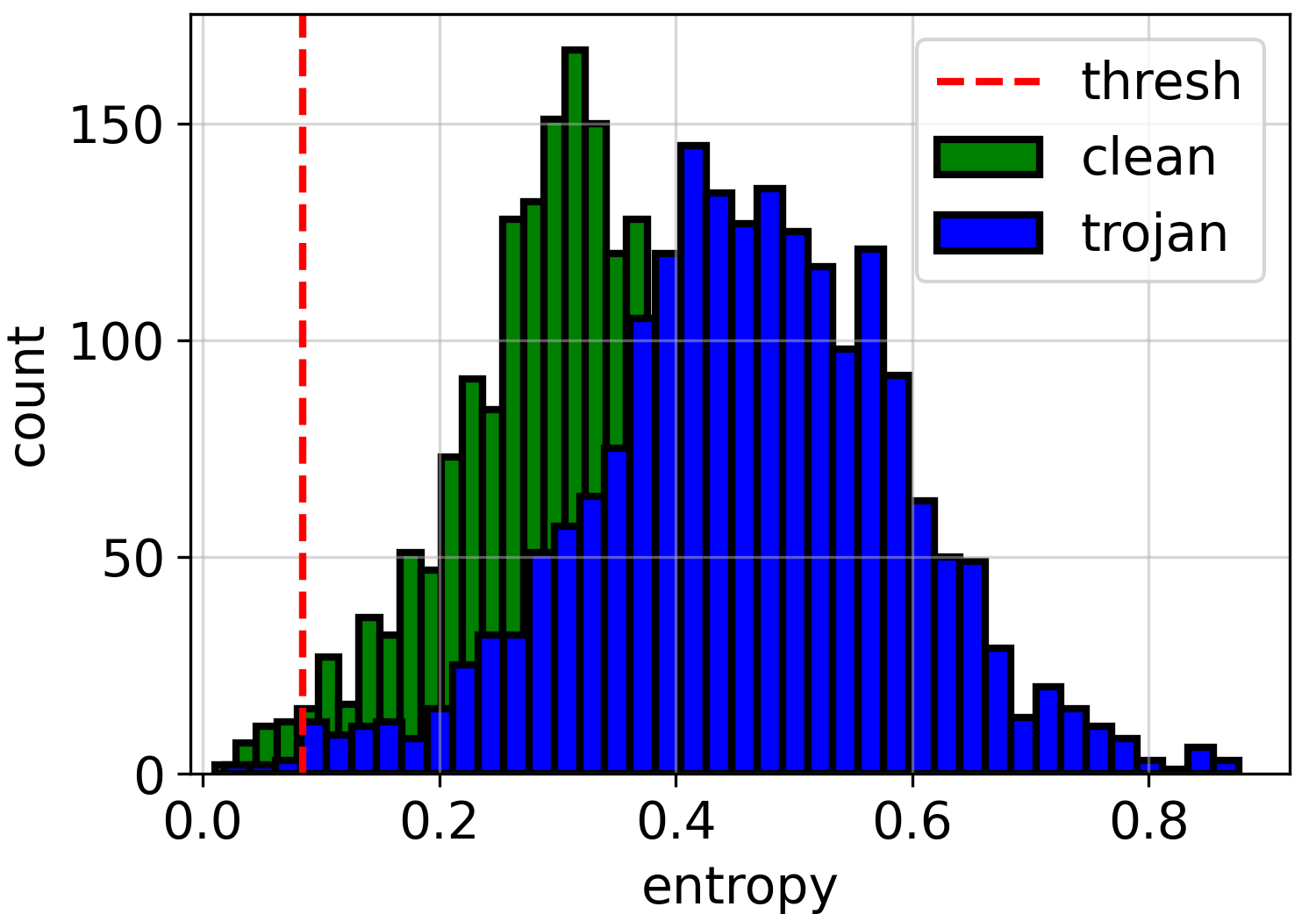}
\par\end{centering}
}~~~\subfloat[Fine Pruning\label{fp}]{\begin{centering}
\includegraphics[width=0.4\textwidth]{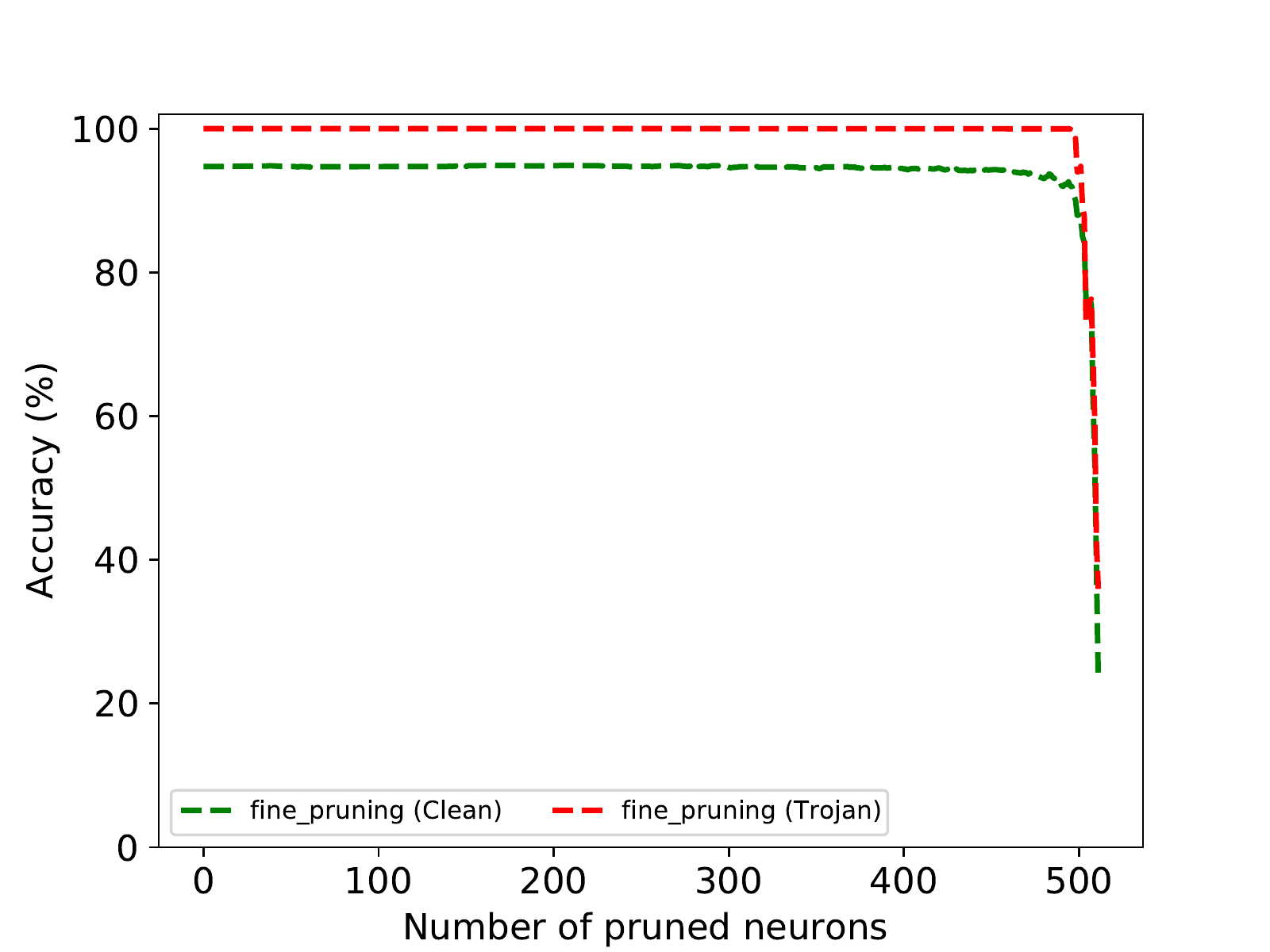}
\par\end{centering}
}\caption{a) STRIP results of CIFAR10 on pure and Trojan data on a target class
(class \emph{6}). The Trojan data for each target class is created
with their corresponding class-specific triggers. b) Fine Pruning
on MTTA CIFAR10 8x8 Trojan model.\label{sota_defense}}
\end{figure*}

Figure \ref{strip-2} shows the entropy plots of pure images and Trojan
images for a target class (class \emph{6})\emph{ }after performing
STRIP\emph{. }The remaining entropy plots and False Positive Rate
(FPR)\emph{, }and False Negative Rate (FNR) is reported in supplementary\emph{.
}The threshold of the entropy is calculated from the pure images
by assuming that it follows a normal distribution. The 1\% of the
normal distribution of the entropy of the pure images will be chosen
as the threshold to separate pure and Trojan images. So, during test
time, any inputs which have an entropy value above the threshold will
be considered as a pure image. From the Figure \ref{strip-2}, it
is evident that it is difficult to separate the pure and Trojan images
with the computed threshold (shown as a red dotted vertical line).
The results show that STRIP totally fails to defend against MTTA attacks.

Fine Pruning \cite{liu2018fine} results of the CIFAR-10 8$\times$8
trigger trained MTTA model is shown in Figure \ref{fp}. This mechanism
removes the least activated neurons based on the pure images the defender
has access to. Thus, this mechanism will prune the neurons which are
highly influenced by the Trojan features and drops the Trojan accuracy
of the model. It is clear that the Trojan and pure accuracy remains
intact as the pruning progresses and drops together when the number
of pruned neurons is equal to the total number of neurons.

For STS \cite{harikumar2021scalable} we performed the experiments
on CIFAR10 8$\times$8 trigger trained MTTA model we find that Trojan
model is detected as Trojan with one class detected as the target
class. However we have noticed that a pure CIFAR10 model is also detected
as Trojan. This expose the fact that for a large and complex network
it is possible to find a generalizable trigger for a pure model. 

Neural Attention Distillation \cite{li_21neural} uses attention based
knowledge distillation to fine tune a student model from the given
Trojan model. We have done experiments on results on the CIFAR10 8$\times$8
trigger trained MTTA model which gives a pure accuracy of $90.01$,
and Trojan accuracy drops to $55.09$ percentage. NAD performs partially
well in reducing the Trojan accuracy with only $4$ percentage reduction
in the pure accuracy. However, the performance of the model in detecting
the true class of the Trojan data is only $47.7$ percentage. This
shows that the even though it reduces the Trojan accuracy it is still
not classifying the Trojan inputs to it actual class.

\subsection{Invisibility of MTTA}

Here, we test the invisibility of the MTTA attack for pure models.
We use a pure model to check for the pure accuracy for images under
all three attacks: MTTA, Badnet and input-aware attack. It is clear
from the Table \ref{invi_pure} that both the Badnet and MTTA attacks
are invisible to the pure model. However, there is a slight drop (>1\%)
of performance for input-aware attacks. This is expected because the
amount of changes for input-aware attacks are much more than the trigger
based attacks by MTTA and Badnet and thus pure accuracy is affected.
Even a slight drop in pure accuracy is enough to make it vulnerable
to early detection.
\begin{table*}
\centering{}\caption{Pure accuracy for Trojan data by a Pure model on both 4x4 and 8x8
triggers of CIFAR10 dataset.\label{invi_pure}}
\begin{tabular}{|c|c|c|c|c|c|}
\hline 
\multirow{3}{*}{Pure accuracy} & \multicolumn{5}{c|}{Pure model accuracy on Trojan data}\tabularnewline
\cline{2-6} \cline{3-6} \cline{4-6} \cline{5-6} \cline{6-6} 
 & \multicolumn{2}{c|}{MTTA} & \multicolumn{2}{c|}{Badnet} & \multirow{2}{*}{Input-aware attack}\tabularnewline
\cline{2-5} \cline{3-5} \cline{4-5} \cline{5-5} 
 & 4$\times$4 & 8$\times$8 & 4$\times$4 & 8$\times$8 & \tabularnewline
\hline 
94.55 & 94.54 & 94.54 & 94.54 & 94.54 & \emph{93.41}\tabularnewline
\hline 
\end{tabular}
\end{table*}

\subsection{Trojan Detection}

We look at both the class-wise detection and model level detection
performance. A class is declared Trojan if any of the recovered triggers
for that class produces a class-distribution entropy that is lower
than the threshold (as per Eq \ref{eq:threshold}). We use $\delta=0.1$
in all our experiments. A model is flagged as Trojan when at least
one of the classes is Trojan.
\begin{table*}
\centering{}\caption{a) F1-score for class-wise detection. b) F1-score for Trojan model
detection with 90\% Trojan effectiveness for MTD and threshold \emph{2.0}
for NC.}
\subfloat[Class detection.\label{f1-score-class-detectn}]{%
\begin{tabular}{|c|c|c|c|c|}
\hline 
\multirow{3}{*}{Dataset} & \multicolumn{4}{c|}{F1-score target class detection}\tabularnewline
\cline{2-5} \cline{3-5} \cline{4-5} \cline{5-5} 
 & \multicolumn{2}{c|}{4$\times$4} & \multicolumn{2}{c|}{8$\times$8}\tabularnewline
\cline{2-5} \cline{3-5} \cline{4-5} \cline{5-5} 
 & NC & MTD & NC & MTD\tabularnewline
\hline 
MNIST & 0.0 & \textbf{0.92} & 0.0 & \textbf{1.0}\tabularnewline
\hline 
GTSRB & 0.0 & \textbf{0.81} & 0.0 & \textbf{0.77}\tabularnewline
\hline 
CIFAR10 & 0.0 & \textbf{1.0} & 0.0 & \textbf{1.0}\tabularnewline
\hline 
YTF & 0.0 & \textbf{0.43} & 0.0 & \textbf{0.46}\tabularnewline
\hline 
\end{tabular}}~~~~\subfloat[Model detection.\label{tab:F1-score-model-detection}]{%
\begin{tabular}{|c|c|c|}
\hline 
\multirow{2}{*}{Dataset} & \multicolumn{2}{c|}{F1-score model detection}\tabularnewline
\cline{2-3} \cline{3-3} 
 & NC & MTD\tabularnewline
\hline 
MNIST & 0.0 & \textbf{1.0}\tabularnewline
\hline 
GTSRB & 0.0 & \textbf{1.0}\tabularnewline
\hline 
CIFAR10 & 0.0 & \textbf{1.0}\tabularnewline
\hline 
YTF & 0.0 & \textbf{1.0}\tabularnewline
\hline 
\end{tabular}}
\end{table*}

\subsubsection{Model Detection}

We have a pure model and two Trojan models corresponding to two different
trigger sizes for each dataset. The F1-score of the model detection
by Neural Cleanse (NC) and our method is shown in Table \ref{tab:F1-score-model-detection}.
It is clear from the Table that NC failed to detect Trojan in the
MTTA setting. It is also interesting to note that Pure models of all
the datasets are getting detected as Trojan models in NC. However,
our proposed detection mechanism MTD has an F1-score of 1.0 for all
the datasets clearly separating Trojans from Pure.

\subsubsection{Class-wise Detection}

In Table \ref{f1-score-class-detectn} we report the F1-score of NC
and MTD in detecting the target classes. For MNIST and CIFAR-10 it
was able to detect the target classes correctly. However for GTSRB
and YTF there has been a drop in the class-wise detection performance.
The drop happens because in those datasets (traffic signs, and faces)
many of the classes are quite similar and when among a group of similar
classes one is target class then we observe that many of the others
also happen to be detected as target class as well (detail illustration
is provided in the supplementary). This is expected because the shortcut
introduced by a target class also end up serving the classes close
by. As expected, NC failed badly because it was not designed to detect
multi-target attack.
\begin{figure*}
\begin{raggedright}
\subfloat[Original images ($x$)\label{fig:Original-images_nontro}]{\includegraphics[width=0.28\textwidth]{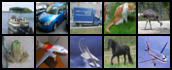}

}~~\subfloat[Optimised images ($x'$)\label{fig:Optimized-images_nontro}]{\includegraphics[width=0.28\textwidth]{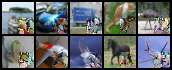}

}~~\subfloat[$\Delta x_{rev}=x'-x$\label{fig:delx_nontro}]{\includegraphics[width=0.28\textwidth]{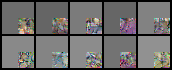}

}\\
\subfloat[Original images ($x$).]{\includegraphics[width=0.28\textwidth]{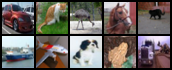}

}~~\subfloat[Optimised images ($x'$).]{\includegraphics[width=0.28\textwidth]{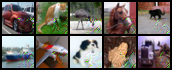}

}~~\subfloat[$\Delta x_{rev}=x'-x$]{\includegraphics[width=0.28\textwidth]{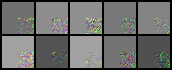}}~~\subfloat[Original trigger\label{fig:Original-trigger}]{\includegraphics[width=0.17\textwidth]{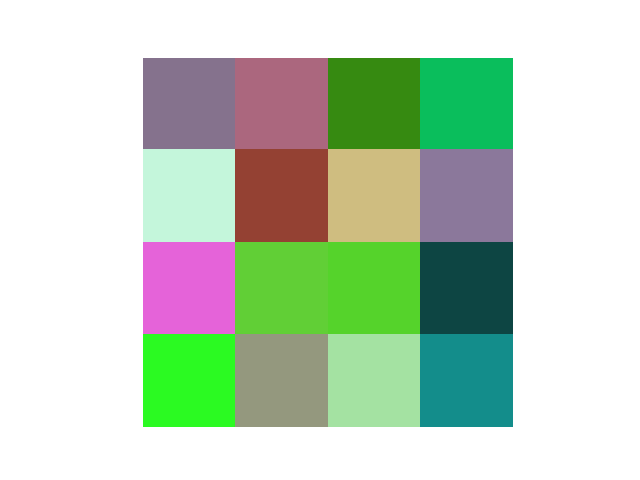}}
\par\end{raggedright}
\caption{Sample reverse-engineered triggers for non-Trojan class (top row)
and Trojan class (bottom row) of a CIFAR-10 Trojan model with $4\times4$
trigger. Please zoom in to see how the non-Trojan perturbations are
optimising more towards part of the image of the target class whilst
the Trojan triggers are optimising towards the original trigger. \label{fig:sample_non-trojan_trojan}}
\end{figure*}

The initial set of images, the optimised images, the difference between
the given images and the optimised images ($\Delta x_{rev}$), of
a non-target class (\emph{class 7}) and a target class (\emph{class
6}) is shown in the top and bottom rows of Figure \ref{fig:sample_non-trojan_trojan},
respectively for a CIFAR-10 Trojan model trained with $4\times4$
trigger. The $\Delta x_{rev}$ of the non-Trojan class samples have
no visible trigger patterns in it, however, for the Trojan class there
are some patterns which look like the original trigger as shown in
Figure \ref{fig:Original-trigger}. More samples of non-target and
target classes for all datasets are reported in the supplementary.

\subsubsection{Against adaptive attack}

We can consider the defense against a simple adaptive attack scenario.
Assume a situation where we can divide the data distribution of a
class into two or more separate sub-distributions, where only one
sub-distribution is triggered, but not others. We think such a division
would be very unlikely to be achieved. For example, it would be very
hard, if not impossible, to separate STOP sign class (as in GTSRB
dataset) into two very distinct sub-distributions such that a trigger
works for only one kind of STOP sign images. If though such can be
done (e.f., for the CAT class only black cats are triggered), then
our detection mechanism may fail. Especially so, if the triggered
sub-distribution is only a tiny part of the whole distribution. However,
that also means that opportunity to deploy backdoor successfully is
diminished. In conclusion, while we think such an adaptive attack
can defeat our MTD, they are neither easy or feasible in all scenarios.

To demonstrate the efficiency of the MTTA attack we have used only
20\% of the total training data from CIFAR-10 to train an 8$\times$8
trigger trained MTTA model. We have found that the Trojan effectiveness
is still $99.94$ however the pure accuracy of the model drop significantly
by $10$ percentage.
\begin{table*}
\centering{}\caption{The performance of MTD mechanism on MTTA Trojan models trained on
different datasets.}
\begin{tabular}{|c|c|c|c|c|c|c|c|c|c|c|}
\hline 
\multirow{2}{*}{Dataset} & \multicolumn{10}{c|}{F1-score}\tabularnewline
\cline{2-11} \cline{3-11} \cline{4-11} \cline{5-11} \cline{6-11} \cline{7-11} \cline{8-11} \cline{9-11} \cline{10-11} \cline{11-11} 
 & \multicolumn{5}{c|}{4$\times$4 trigger} & \multicolumn{5}{c|}{8$\times$8 trigger}\tabularnewline
\hline 
\emph{$\delta$} & \emph{0.01} & \emph{0.05} & \emph{0.1} & \emph{0.15} & \emph{0.20} & \emph{0.01} & \emph{0.05} & \emph{0.1} & \emph{0.15} & \emph{0.20}\tabularnewline
\hline 
MNIST & \textbf{0.92} & 0.82 & 0.82 & 0.82 & 0.82 & \textbf{1.0} & 1.0 & 0.93 & 0.93 & 0.93\tabularnewline
\hline 
GTSRB & 0.28 & 0.66 & \textbf{0.81} & 0.82 & 0.82 & 0.56 & 0.61 & \textbf{0.77} & 0.82 & 0.82\tabularnewline
\hline 
CIFAR10 & \textbf{1.0} & 1.0 & 1.0 & 1.0 & 0.93 & \textbf{1.0} & 1.0 & 1.0 & 1.0 & 0.93\tabularnewline
\hline 
YTF & 0.02 & 0.28 & \textbf{0.43} & 0.46 & 0.45 & 0.12 & 0.37 & \textbf{0.46} & 0.45 & 0.45\tabularnewline
\hline 
\end{tabular}
\end{table*}

\subsection{Ablation Study}

\subsubsection{Performance Vs $\delta$}

We report the F1-score of Trojan class detection for different Trojan
models based on different values of $\delta$. The results shows that
as we increase $\delta$, the F1-score reduces. This is because with
higher $\delta$ many non-Trojan classes are also classified as Trojan
classes. We find that $\delta=0.1$ provides the most stable results
across all the datasets.

\subsubsection{Performance with and without mask}

We use the 8$\times$8 triggers trained CIFAR10 MTTA model to perform
experiments with and without mask.
\begin{table}
\centering{}\caption{F1-score model and class detection (with 90\% Trojan effectiveness).\label{tab:f1_wd_wo_mask}}
\begin{tabular}{|>{\raggedright}p{0.15\columnwidth}|>{\centering}p{0.15\columnwidth}|>{\centering}p{0.15\columnwidth}|>{\centering}p{0.15\columnwidth}|>{\centering}p{0.15\columnwidth}|}
\hline 
\multirow{2}{0.15\columnwidth}{Dataset} & \multicolumn{2}{>{\centering}p{0.3\columnwidth}|}{F1-score model detection} & \multicolumn{2}{>{\centering}p{0.3\columnwidth}|}{F1-score class detection}\tabularnewline
\cline{2-5} \cline{3-5} \cline{4-5} \cline{5-5} 
 & 8$\times$8 with mask & 8$\times$8 without mask & 8$\times$8 with mask & 8$\times$8 without mask\tabularnewline
\hline 
CIFAR10 & 1.0 & 0.0 & 1.0 & 0.0\tabularnewline
\hline 
\end{tabular}
\end{table}
\begin{figure*}
\centering{}\includegraphics[width=0.3\textwidth]{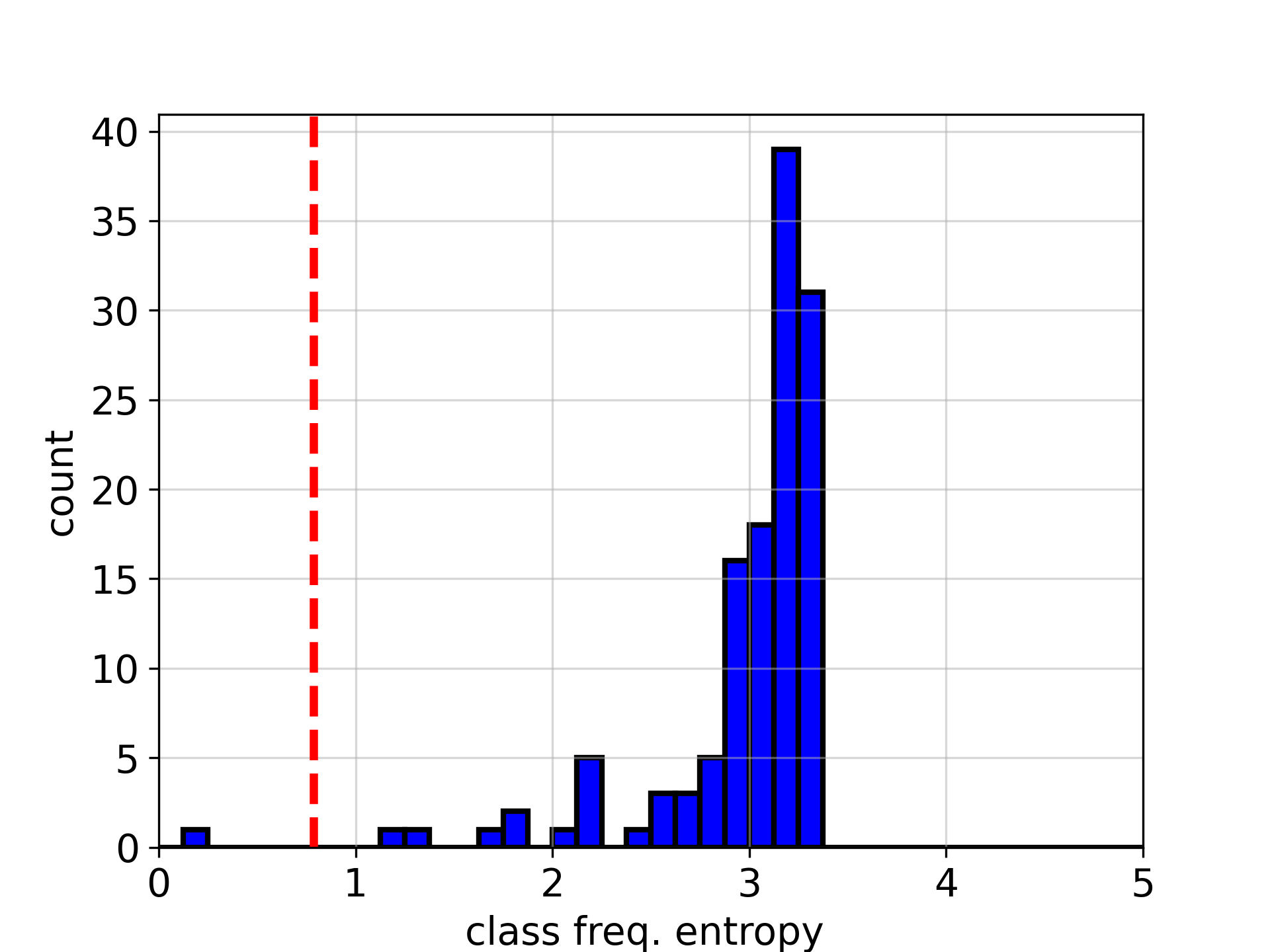}~~\includegraphics[width=0.3\textwidth]{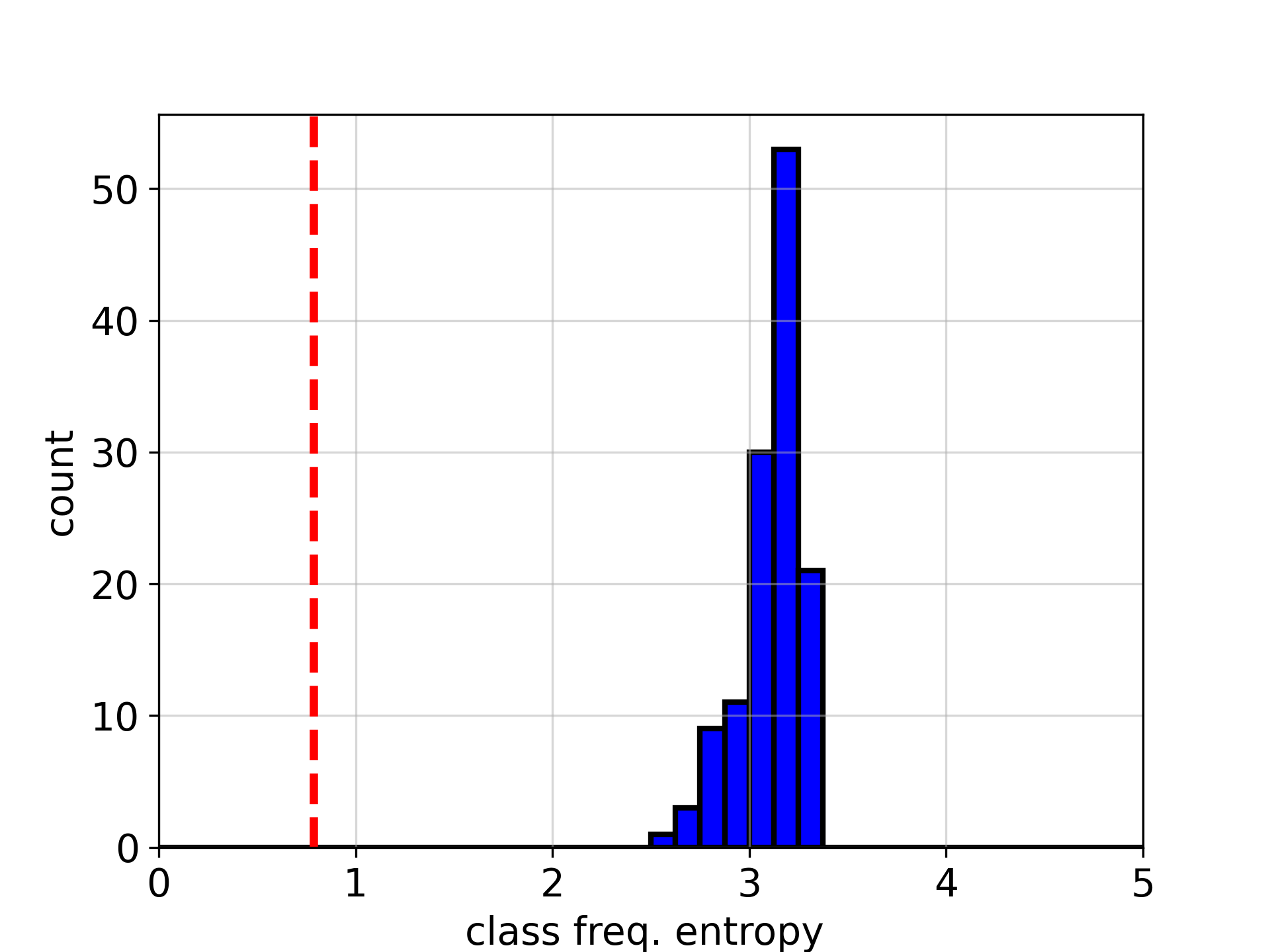}~~\includegraphics[width=0.3\textwidth]{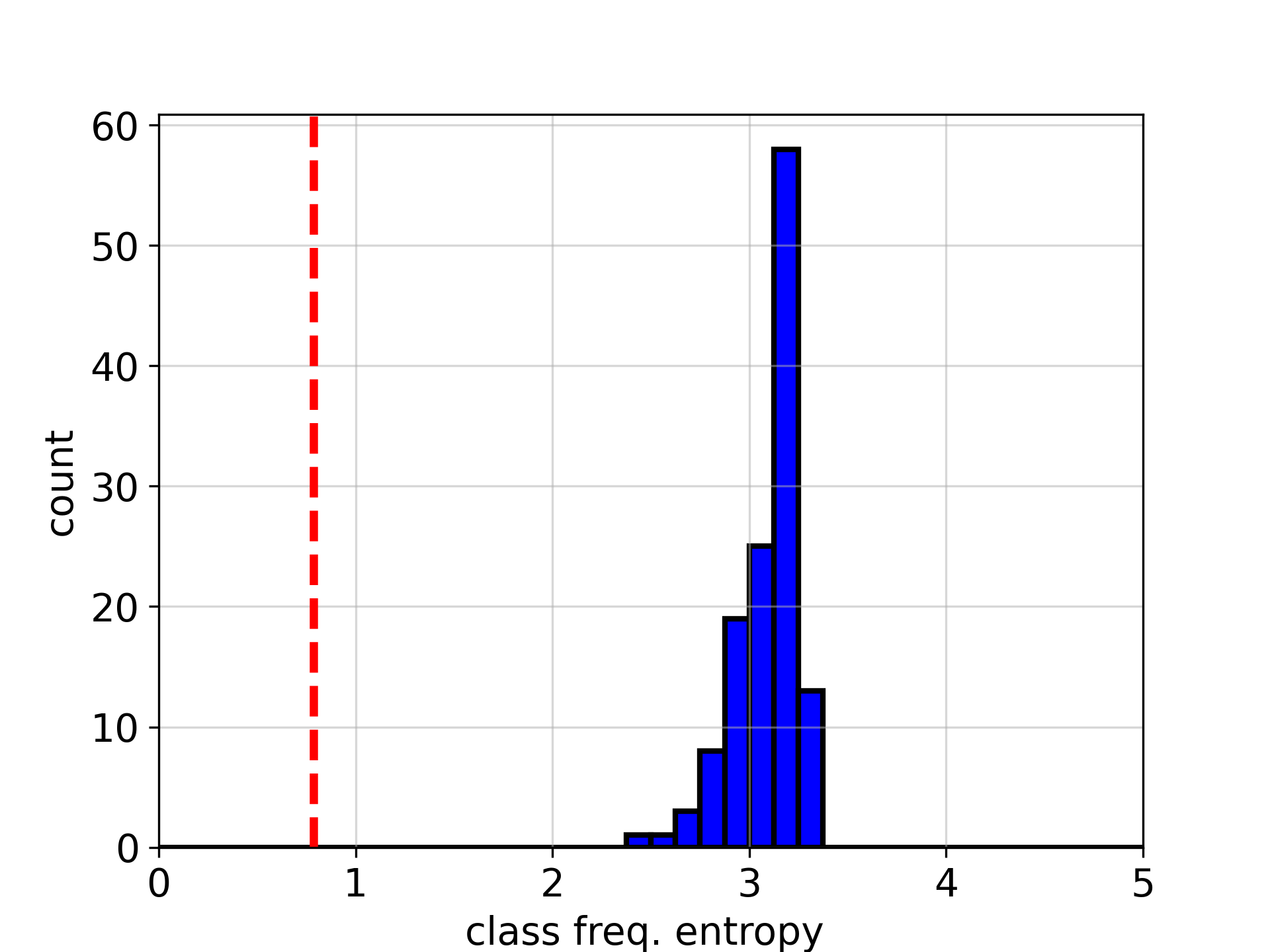}\caption{Distribution of class distribution entropies computed over many recovered
triggers for both Trojan (left) and non-Trojan classes (middle, right)
for single target Badnet attack. Only the Trojan class has some triggers
that resulted in entropy scores lower than the threshold (red dashed
line).\label{fig:single_tgt_setting}}
\end{figure*}
 The Table \ref{tab:f1_wd_wo_mask} shows that for without mask our
method won't be able to detect the Trojan classes and hence the Trojan
model. However, it achieves a perfect F1-score when used with mask.
This shows the importance of using mask in the MTD. 

\subsubsection{Single target attack}

We choose a Badnet trained on CIFAR-10 dataset with 4x4 trigger and
apply our MTD. For the Badnet, \emph{class 0 }is the target class
and the rest are non-target classes. When MTD is applied, only the
target class is detected as Trojan and all the non-target classes
are detected as non-Trojan. The entropy plots which is shown in Figure
\ref{fig:single_tgt_setting} of the Trojan (\emph{class 0}) and a
randomly sampled two non-Trojan classes (\emph{class 2 }and\emph{
class }9) demonstrate the difference between the entropy distributions.
Here we assumed that we know the location of the trigger. Even if
the location is not known we can use MTD by placing our mask across
the image and performing 9 (4 quadrants + 4 at the intersection on
the quadrants + 1 in the middle) trigger reverse engineering optimisation.

%% file: limitation.tex
Our present work has the following limitations:
\begin{itemize}
\item We tested the efficacy of the attacks on fixed image datasets only.
In physical domain our proposed attack may become less robust due
to the presence of environmental disturbances. However, that will
affect all attack methods. In relative terms, MTTA may still provide
a better attack model especially when the attack is carried out in
physical space. When environmental conditions are more favorable e.g.,
in clear daylight, Trojan attacks would work quite well and thus they
still pose a significant threat.
\item We assumed that the trigger is no more than the size of 1/4th of the
image. One can easily violate this assumption, at the expense of stealth.
In that case, MTD may only find part of the trigger and thus detection
performance may suffer. However, security is a cat and mouse game
between an attacker and a defender. From the defender side our aim
is to make the job of the attacker as hard as possible by limiting
his choice, and we believe we achieved that through this work. 
\item We assume the availability of a detection dataset contains sufficient
number of pure images. However, we did not investigate the cases when
such pure dataset is not available or when purity cannot be guaranteed. 
\end{itemize}

%% file: conclusion.tex
In this paper we proposed a variation of the Badnet style attack on
multiple targets that is able to defeat state-of-the-art defense mechanisms
and are robust than many recent attacks. We then proposed a new detection
method based on reverse-engineering of triggers for individual images
and then verifying if a recovered trigger is transferable. We then
propose a mechanism to compute threshold that would separate the Trojan
triggers from the other triggers based on the class-distribution entropy.
Our extensive experiments on four image datasets of varying number
of classes and dataset size show that we can classify pure and Trojan
models with a perfect score.